%% file: main.tex
\documentclass[runningheads]{llncs}
\usepackage[T1]{fontenc}
\usepackage{graphicx}
\usepackage{multirow}
\usepackage{booktabs}
\usepackage{amsmath}
\usepackage{amssymb}
\usepackage{subcaption}
\usepackage[misc]{ifsym}
\newcommand{\corr}{(\Letter)}
\usepackage{enumitem}
\usepackage{xcolor}

\usepackage{amsthm}
\usepackage{bm}
\usepackage{mathtools}
\usepackage[english]{babel}
\usepackage{amsxtra}
\usepackage{algorithm}
\usepackage{algpseudocode}
\usepackage{lipsum}
\usepackage{amsfonts}
\usepackage{subfig}
\usepackage{url}
\usepackage{epstopdf}
\usepackage{caption}
\usepackage{float}

\newcounter{ManosNOC}
\stepcounter{ManosNOC}

\newcounter{ProfessorNOC}
\stepcounter{ProfessorNOC}

\newcounter{AntonisNOC}
\stepcounter{AntonisNOC}

\newcounter{DusanNOC}
\stepcounter{DusanNOC}

\begin{document}

\title{Robust Federated Learning under Adversarial Attacks via Loss-Based Client Clustering}

\titlerunning{Robust FL via Loss-Based Client Clustering}

\author{Emmanouil Kritharakis\inst{1} \corr \and Dusan Jakovetic\inst{2} \and Antonios Makris\inst{1} \and
Konstantinos Tserpes\inst{1}}

\authorrunning{E. Kritharakis et al.}

\institute{School of Electrical and Computer Engineering, National Technical University of Athens, Athens, Greece \\ \email{\{ekritharakis,antoniosmakris,tserpes\}@mail.ntua.gr}
\and
Faculty of Sciences, University of Novi Sad, Novi Sad, Serbia,
\email{dusan.jakovetic@dmi.uns.ac.rs}}

\maketitle

\begin{abstract}
Federated Learning (FL) enables collaborative model training across multiple clients without sharing private data. We consider FL scenarios wherein FL clients are subject to adversarial (Byzantine) attacks, while the FL server is trusted (honest) and has a trustworthy side dataset. This may correspond to, e.g., cases where the server possesses trusted data prior to federation, or to the presence of a trusted client that temporarily assumes the server role. Our approach requires only two honest participants, i.e., the server and one client, to function effectively, without prior knowledge of the number of malicious clients. Theoretical analysis demonstrates bounded optimality gaps even under strong Byzantine attacks. Experimental results show that our algorithm significantly outperforms standard and robust FL baselines such as Mean, Trimmed Mean, Median, Krum, and Multi-Krum under various attack strategies including label flipping, sign flipping, and Gaussian noise addition across MNIST, FMNIST, and CIFAR-10 benchmarks using the Flower framework.

\end{abstract}

\input{Introduction}
\input{Related_Work}
\input{Problem_Formulation}
\input{Experiments}
\input{Conclusion}
\input{Acknowledgements}
\bibliographystyle{splncs04}
\bibliography{Bibliography}
\end{document}

%% file: Introduction.tex
\section{Introduction}

We investigate a Federated Learning (FL) setting involving a group of clients, each possessing a distinct partition of a shared dataset. The overarching goal is to collaboratively train a global model on the union of all data partitions without requiring clients to share their private data, a scenario where FL is particularly well suited due to its privacy-preserving nature \cite{hu2024overview,li2020federated}. 

In such a setting, while client-to-client communication is permitted, not all participants can be assumed to act honestly. A subset of clients, unspecified in size and distribution, may behave maliciously by attempting to introduce bias or poison the global model in pursuit of adversarial objectives. This scenario presents the challenge of designing a robust mechanism that can mitigate the impact of such adversaries, ensuring convergence to an optimal model accuracy (as if no attackers were present), subject to an acceptable degradation in accuracy and convergence rate \cite{hu2024overview,xia2023poisoning}.

To address this challenge, we propose a robust FL algorithm that assumes a trusted server with access to a side-trusted dataset. This scenario may correspond to, e.g., an honest client taking the role of the FL server, or to scenarios where the server has a trusted dataset prior to federation with the clients. This server maintains a global model by aggregating the clients' local models; for this, the server makes use of its local trusted dataset to form an empirical loss function $f_{\mathrm{S}}$ that is a proxy to model quality. In more detail, the server ranks the clients' local models based on the $f_{\mathrm{S}}$ values evaluated at their local models and clusters the clients into two groups accordingly, those estimated to be honest and those estimated to be malicious (subject to an attack). We hypothesize that clustering clients based on their respective $f_{\mathrm{S}}$ loss values allow us to effectively isolate malicious contributions and optimize convergence and performance.

Our approach relies on the following premises: (i) at least two clients are honest, one of whom can securely act as the server, being agnostic to which clients are honest or malicious; and (ii) all clients participate in each training round, ensuring that both honest and adversarial behaviors are represented in every aggregation cycle.

We provide a formal mathematical description of the proposed mechanism and the client-clustering algorithm, and validate our hypothesis under the stated assumptions. Experimental evaluations compare our approach against standard and robust aggregation baselines, including Mean, Trimmed Mean, Median, Krum, and Multi-Krum. Simulations are conducted using the Flower framework\footnote{\url{https://flower.ai/}} with 10 clients collaboratively training models on CIFAR-10, MNIST, and Fashion-MNIST (FMNIST) datasets.

Our experimental results demonstrate that the proposed method consistently outperforms existing baselines in terms of centralized accuracy and convergence robustness, even under ongoing attack scenarios where the number of malicious clients is not known a priori. This indicates strong resilience and adaptability under adversarial conditions. The threat models considered include Label Flipping, Sign Flipping, and Noise Addition.

The remainder of the paper is organized as follows: Section \ref{related_work} reviews related work on attack models and defense mechanisms in FL. Section \ref{problem_setting} details our problem formulation and proposed methodology. Section \ref{experiments} presents our experimental results and analysis. Section \ref{conclusions} concludes the paper with a discussion of our findings and future directions.

%% file: Related_Work.tex
\section{Related Work}
\label{related_work}
\subsection{Threat Models}
\label{Threat Models}

Although FL was initially introduced in 2017 \cite{mcmahan2017communication} as a paradigm to improve privacy and security by eliminating the need to transmit raw training data from clients to a central server, this novel approach has simultaneously redefined the landscape of potential attack vectors within distributed machine learning environments \cite{hu2024overview}. Among the various adversarial threats in FL, poisoning attacks, which aim to hinder or mislead the convergence of the global model by injecting malicious updates or manipulating local training data, pose a significant challenge. 

In this work, we focus xon one targeted data poisoning attack, known as Label-Flipping, and two untargeted model poisoning attacks, commonly referred to as Byzantine attacks: namely, Sign-Flipping and Gaussian noise injection. In general, a targeted attack refers to a malicious attempt to disrupt, manipulate, or degrade the performance of the training process or the models of specific participants, whereas an untargeted attack broadly targets the entire system without focusing on particular participants, models, or tasks. The behavior of Byzantine clients can be formally characterized as follows:
Let \( \Delta w_{t+1}^{(i)} \) denote the local model update from client \( i \) at training round \( t \), and let \( \mathcal{B} \) be the set of Byzantine (malicious) clients. The update actually submitted to the server, \( \hat{\Delta w}_{t+1}^{(i)} \), is defined as:

\[
\hat{\Delta w}_{t+1}^{(i)} =
\begin{cases}
\Delta w_{t+1}^{(i)} & ,\text{if } i \notin \mathcal{B} \quad \text{(honest client)} \\
* & ,\text{if } i \in \mathcal{B} \quad \text{(Byzantine client)}
\end{cases}
\]
where $*$ represents an arbitrary value.

Byzantine attacks can be further categorized on the basis of the type of information they exploit. On one hand, attacks such as A Little is Enough (ALIE) \cite{baruch2019little} relies on access to malicious clients in order to craft statistically coordinated adversarial updates, while Inner Product Manipulation (IPM) \cite{xie2020fall} compromises the aggregation process by replacing a subset of client updates on the server side with carefully designed Byzantine gradients aimed at maximizing their disruptive impact. On the other hand, some attacks operate independently, relying solely on the local training data and the model updates computed by each individual malicious client during each federated learning round. In our work, we assume that malicious clients act independently, without access to or use of information from other clients to enhance their attack strategies. Such representative state-of-the-art poisoning attacks are described as follows.

\textbf{Label Flipping} \cite{guo2021siren,li2023experimental}: Malicious clients alter their local training data by mislabeling class labels prior to local model training. The objective is to compromise the integrity of the global model by injecting structured label inconsistencies that lead to degraded performance. Formally, let $C$ denote the total number of classes and $c$ the original class label. Under a label flipping attack, each label $c$ is systematically assigned to a new label given by $C-c-1$.

\textbf{Sign Flipping} \cite{guo2021siren,li2023experimental}: Byzantine clients adversarially alter their local model updates prior to transmission to the server. In particular, they invert the sign of each element in the gradient vector, thereby reversing the direction of the update and impairing the convergence of the global model.

\textbf{Noise Addition} \cite{fang2020local,li2023experimental}: Byzantine clients perturb their local model updates by injecting random noise prior to the server-side aggregation. Let 
$\Delta w$ denote the genuine local model update; instead of transmitting $\Delta w$, a Byzantine client submits 
$\Delta w + \epsilon$, where $\epsilon \sim \mathcal{N}(\mu,\sigma^2\mathcal{I})$ represents additive Gaussian noise with mean value $\mu$ and variance $\sigma^2\mathcal{I}$. This strategy aims to degrade the integrity of the global model by increasing the variance in the aggregated update. In our experimental setup involving the Noise Addition attack, the injected noise follows a distribution with a mean of $\mu = 0.25$ and a fixed variance of $\sigma^2 = 1$.

\subsection{Resilient Aggregation Methods Against Byzantine Behavior}
\label{Resilient Aggregation Methods Against Byzantine Behavior}

Federated Learning is susceptible to poisoning attacks, in which adversarial participants deliberately send misleading or malicious updates to the central server, undermining the integrity of the global model. To address the threat of adversarial attacks within FL, it is essential to adopt robust aggregation methods designed to be resilient against such malicious behavior. Robust aggregators incorporate strategies to detect and exclude malicious inputs from malicious participants. These methods typically involve techniques such as identifying outliers, applying robust statistical measures, and setting thresholds to filter out harmful updates, thus maintaining the integrity of the global model even when adversaries are present. Numerous aggregation operators based on robust estimators have been proposed; we highlight the following ones:

\textbf{Mean} \cite{mcmahan2017communication}: Serves as the standard aggregation technique in federated learning, known as FedAvg, where the central server computes a data-size-weighted element-wise average of the model updates received from participating clients. Each client’s contribution is proportional to the number of local data samples it holds, ensuring that the global update reflects the underlying data distribution. Under the assumptions of independently and identically distributed (IID) data and honest client participation, this method is computationally efficient. However, it lacks robustness against adversarial inputs: a single malicious client can significantly distort the average due to its sensitivity to outliers and lack of robustness guarantees \cite{lu2024federated}, \cite{tang2022fedcor}. This vulnerability renders Mean aggregation susceptible to various poisoning attacks, including sign flipping, label flipping, and noise injection strategies. As such, it is often used as a baseline for benchmarking more robust aggregation rules designed to mitigate adversarial influence.

\textbf{Trimmed Mean} \cite{yin2018byzantine}: Serves as an alternative to standard mean aggregation, employing a coordinate-wise aggregation rule. For each dimension, the set of model updates among all clients is sorted, and a fraction 
\( \beta \in [0, 0.5) \)  of the lowest and highest values is removed, effectively trimming both tails of the distribution. The mean is then computed over the remaining central values. This approach mitigates the influence of extreme or potentially malicious inputs, making it suitable for adversarial FL use cases.

\textbf{Median} \cite{yin2018byzantine}: Another coordinate-wise aggregation rule, following the steps of Trimmed Mean. For each model parameter (i.e., each dimension of the gradient vector), it computes the median of the corresponding values received from all clients. 

\textbf{Krum} \cite{blanchard2017machine}: Krum operates by evaluating each client's model update based on its consistency with those of other clients. Specifically, for each update, it computes the Krum score, known as the sum of squared Euclidean distances to its \( N -f - 2 \) nearest neighboring updates, where \(f\) denotes the expected maximum number of malicious clients. The update with the smallest Krum score—indicating maximal alignment with the majority—is selected for aggregation. This approach enables Krum to effectively suppress outliers and maintain the integrity of the global model in adversarial settings. 

\textbf{Multi-Krum} \cite{blanchard2017machine}: Multi-Krum extends the Krum aggregation rule by selecting multiple client updates rather than a single one, thereby improving convergence stability while maintaining Byzantine robustness. In particular, after calculating the Krum score, it selects the \( k \leq N - f - 2 \) updates with the lowest Krum scores and aggregates them via coordinate-wise averaging. Unlike other defense mechanisms, both Krum and Multi-Krum necessitate prior knowledge of the maximum number of malicious clients within the federated learning setup.

{\setlength{\parskip}{4pt}
\textbf{Bulyan} \cite{guerraoui2018hidden}. Bulyan is a two-phase robust aggregation algorithm developed to enhance FL's resilience against Byzantine clients. Initially, it iteratively employs the Multi-Krum algorithm to identify a subset of reliable updates by excluding outliers; subsequently, it performs a coordinate-wise trimmed mean on this selected subset to further eliminate extreme values. This combined approach leverages both distance-based filtering and statistical trimming, thereby substantially strengthening robustness and mitigating the influence of adversarial or corrupted client updates. To ensure its effectiveness, Bulyan imposes a stricter constraint on the number of Byzantine clients it can tolerate relative to the total number of clients. 

While the aforementioned defense schemes represent many of the current state-of-the-art approaches, the research community has increasingly shifted focus toward uncovering additional factors that influence robust aggregation in federated learning. Li \cite{li2021byzantine} presents a Byzantine-resilient defense mechanism for federated learning that leverages spatial-temporal analysis of client model updates. The approach integrates round-wise clustering to detect and filter out anomalous updates (spatial analysis) with a temporal consistency module that refines global model updates using a momentum-inspired adjustment based on historical update patterns. FLTrust \cite{cao2020fltrust} enhances Byzantine-robust federated learning by leveraging a small, clean dataset on the server to create a trusted reference update. It computes trust scores based on the alignment of client updates with this reference, adaptively weighting and normalizing them to effectively suppress malicious contributions and improve overall model robustness and convergence. Allouah \cite{allouah2024byzantine} examines how client subsampling and local updates affect the robustness of federated learning under Byzantine attacks. They show that common assumptions—such as full participation or single local steps—can significantly influence the effectiveness of robust aggregation rules. Their analysis reveals that subsampling and local training may unintentionally amplify Byzantine effects, highlighting the need to consider system-level design in developing resilient FL algorithms. In contrast to the aforementioned research efforts, our approach ensures global model convergence even under highly adversarial federated learning settings, where the number of malicious clients equals or exceeds that of honest participants.
}

%% file: Problem_Formulation.tex
\section{Problem Formulation}
\label{problem_setting}
\subsection{Federated Learning Setting}

\begin{figure}[t]
    \centering
    \includegraphics[width=0.85\linewidth]{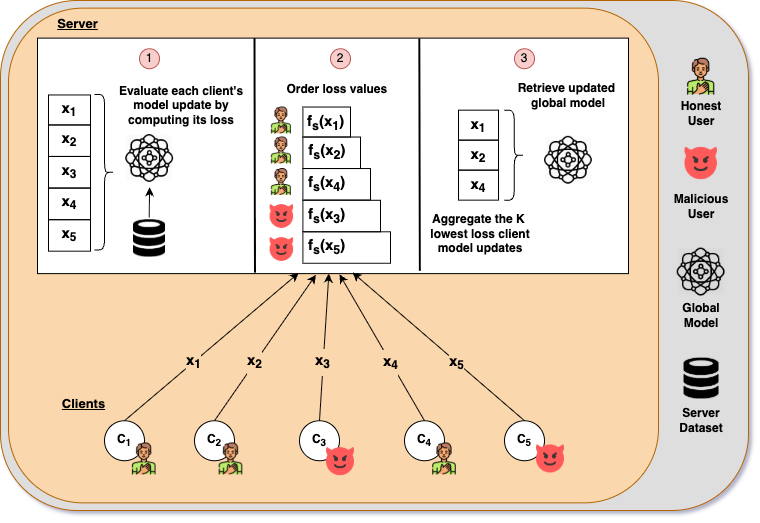}
    \caption{Conceptual view of the proposed loss-based FL defense framework. The server evaluates client submissions using its trusted dataset and aggregates only the most promising models.}
    \label{fig:framework}
\end{figure}

The concept of the proposed solution is presented in Figure \ref{fig:framework}. We consider an FL setting involving $N$ clients and a trusted server, where the goal is to collaboratively minimize the expected population loss:
\begin{equation}
f(x) = \mathbb{E}[F(x, a)].
\end{equation}
where $F: {\mathbb R}^d \times {\mathbb R}^m \rightarrow \mathbb R$
 is the per-data point loss, $x \in \mathbb{R}^d$ denotes the model parameters, and $a \sim \mathcal{P}$, $a \in {\mathbb R}^m$, is a data sample from an unknown distribution $\mathcal{P}$. The operator $\mathbb{E}$ indicates the expectation taken with respect to the distribution $\mathcal{P}$. Each client holds access to a local dataset comprising IID samples drawn from $P$ (or an informative approximation thereof). Specifically, each client $i$ possesses $M_i$ IID data points $b_{i,j}, j=1,...,M_i$, based on which it can form its local empirical loss $f_i(x) = \frac{1}{M_i} \sum_{j=1}^{M_i} F(x,b_{i,j})$. We assume that, during FL training rounds, each client $i$ can query stochastic gradients of $f_i$ for a queried model $x$, e.g., by evaluating a mini-batch stochastic gradient of $f_i$ based on its local data’s random subsample. The server holds another dataset comprised of trusted data points $a_i$, $i=1,...,M_S$, drawn IID from $\mathcal{P}$, where $M_S>0$.

Let the empirical server-side trusted approximation of the population loss be:
\begin{equation}
f_{\mathrm{S}}(x) = \frac{1}{M_S} \sum_{i=1}^{M_S} F(x, a_i).
\end{equation}

The FL training setting proceeds iteratively through communication rounds, indexed by $t \geq 0$, until the global the model converges. At each round, the server broadcasts the current global model $x_t$ to all clients. Each client computes a local update $x_{t+1}^{(i)}$ (e.g., via stochastic gradient
descent (SGD) or Adam) and sends it back to the server. Some clients may act maliciously, submitting poisoned updates $\tilde{x}_{t+1}^{(i)}$ instead.

\subsection{Adversarial Model and Assumptions}

We consider a Byzantine model in which, at each round $t$, an arbitrary (and unknown) subset of clients may act maliciously. The server is assumed to operate in a trusted and secure environment, with no adversarial activity occurring on it during the FL process. Let the assumed attack model at client $i$ and round $t$ be:
\[
\hat{x}_{t+1}^{(i)} = \begin{cases}
x_{t+1}^{(i)}, & \text{if client $i$ is honest}, \\
\tilde{x}_{t+1}^{(i)}, & \text{if client $i$ is malicious}.
\end{cases}
\]
where $\tilde x_{t+1}^{(i)}$ is an arbitrary vector in $\mathbb{R}^d$.
We assume at least one honest client exists per round and that the trusted server can evaluate all received updates with respect to function $f_{\mathrm{S}}(\cdot)$. The function $f_{\mathrm{S}}$ serves as a proxy for the true loss $f$ in (1) and is used to score and rank client updates.

\subsection{Loss-Based Aggregation Mechanism}

The proposed defense strategy follows the steps outlined in Algorithm~\ref{alg:proposed}, where, according to $f_{\mathrm{S}}$, only the top-performing updates are used to update the global model. 

\begin{algorithm}[H]
\caption{Loss-Based Federated Learning with Client Filtering}
\label{alg:proposed}
\begin{algorithmic}[1]
\State Initialize global model $x_0$
\For{each round $t = 0, 1, \dots$}
    \State Server broadcasts $x_t$ to all clients
    \For{each client $i = 1,\dots,N$ \textbf{in parallel}}
        \State Client computes $x_{t+1}^{(i)}$ by performing $R$ steps of local optimizer from $x_t$
        \State Client submits possibly poisoned update $\hat{x}_{t+1}^{(i)}$
    \EndFor
    \State Server evaluates $v_t^{(i)} = f_{\mathrm{S}}(\hat{x}_{t+1}^{(i)})$ for all $i$
    \State Select subset $\mathcal{I}_t$ of $K_t$ clients with lowest $v_t^{(i)}$
    \State Aggregate and update global model: $x_{t+1} = \frac{1}{K_t} \sum_{i \in \mathcal{I}_t} \hat{x}_{t+1}^{(i)}$
\EndFor
\end{algorithmic}
\end{algorithm}

The algorithm begins by arbitrarily initializing the global model parameters, denoted by $x_0$. At the beginning of each FL communication round $t$, the server broadcasts the current global model $x_t$ to all participating clients.
Each client performs local training based on the received global model $x_t$ using a standard optimizer. Let $R \geq 1$ denote the number of local update steps performed at each round. Specifically, the client $i$ computes its updated model $x_{t+1}^{(i)}$ by applying $R$ iterations of an optimization method such as SGD or the Adam optimizer to its local dataset.
The use of multiple local steps reduces communication overhead,  while preserving the locality of the data. Upon completion of local training, both benign and malicious clients concurrently submit their updated model parameters, denoted as 
$\hat{x}_{t+1}^{(i)}$, to the server.

Figure~\ref{fig:framework} provides a step-by-step illustration of the server-side operations of Algorithm~\ref{alg:proposed}. Each submitted model update $\hat{x}_{t+1}^{(i)}$ is evaluated by computing its corresponding loss value $v_t^{(i)} = f_{\mathrm{S}}(\hat{x}_{t+1}^{(i)}) $ using a set of trusted data points $ \{a_i\}_{i=1}^{M_S} $. Let $ K_t $ denote the number of benign model updates received by the server at communication round $ t $. While $ K_t $ may be predetermined in certain defense schemes \cite{blanchard2017machine}, we consider it to be unknown in this work. To estimate $ K_t $, a 2-means clustering algorithm is applied to the set of scalar loss values $ \{v_t^{(i)}\}_{i=1}^N $. The cluster centers are initialized using the minimum and maximum values from this set. The remaining loss values are then assigned to one of the two clusters based on their empirical loss under $ f_{\mathrm{S}} $, resulting in a partition into a low-loss and a high-loss group. The global model is subsequently updated by aggregating only the model updates associated with the low-loss group, whose cardinality corresponds to $ K_t $ in Algorithm~\ref{alg:proposed}.

This approach enables flexible and robust aggregation without requiring prior knowledge of the number of adversaries. The core intuition is that honest clients, being trained on IID or similar data and applying unbiased optimization, tend to yield lower empirical loss on the server’s trusted data.

We advocate that the proposed framework provides a simple yet effective method to mitigate poisoning attacks in FL. It operates under minimal assumptions (only requiring one honest client and a server with a small trusted dataset) and can function without any explicit knowledge of the number of malicious participants, while manipulations over currently attacked clients' models are arbitrary. 

Our theoretical analysis demonstrates that the algorithm achieves bounded errors under appropriately defined metrics. More precisely,
we can establish bounds on the following quantities for all $t \geq 1$: 
1) $\frac{1}{t}\sum_{\ell=0}^{t-1} \mathbb{E}\left[ 
\|\nabla f(x_{\ell})\|^2 \right]$, for non-convex settings; and 
2) $\mathbb{E}\left[ f(x_t) - f^\star\right]$, for convex settings, 
where $f^\star: = \inf_{x \in {\mathbb R}^d}f(x)$ and $\nabla f(x)$ is the gradient of function $f$ at argument $x$.  
 The achieved bounds on these quantities scale predictably with the size and quality of the server-side validation set.

The detailed theoretical analysis, rigorous assumptions, theorems and proofs, are omitted here for brevity. Instead, we provide insight into some key arguments for why Algorithm~1 exhibits robustness to Byzantine attacks. Assume that
\[
\sup_{x \in {\mathbb R}^d} |f_{\mathrm{S}}(x) - f(x)| \leq \delta,
\]
for some quantity $\delta>0$. Under appropriate assumptions, the inequality holds with $\delta= O(1/M_{\mathrm{S}})$ with high probability; see, e.g.,  \cite{BottouEtAl2007,MokhtariEtAl2016}.
 Next, consider a fixed arbitrary training round $t\geq 0$.
 Denote by $\mathcal{N}_t \subseteq  \{1,...,N\}$
 the set of clients that
 is not subject to attacks at round $t$. Also, let $N_t = |\mathcal{N}_t|$ be the set's cardinality. Note that, by our assumptions, $\mathcal{N}_t$ is non-empty. Let the Algorithm 1's parameter $K_t=1$. Then, there holds:
  \begin{equation}
\label{eqn-proof-1}
f_{\mathrm{S}}
\left(x_{t+1}\right)
=
\min_{i \in \{1,...,N\}}
f_{\mathrm{S}}(\hat{x}_{t+1}^{(i)})
\leq
\min_{i \in \mathcal{N}_t}
f_{\mathrm{S}}(x_{t+1}^{(i)})
\leq
\frac{1}{N_t}
\sum_{i \in \mathcal{N}_t} f_{\mathrm{S}}
\left(x^{(i)}_{t+1}\right).
  \end{equation}
  The first inequality in
  \eqref{eqn-proof-1}
  is an important step here that
  essentially ``ceases out'' the role of attacked clients in the analysis.
   Then,
   we have:
   \begin{eqnarray}
 f(x_{t+1})
\leq f_{\mathrm{S}}(x_{t+1}) + \delta
\leq
\frac{1}{N_t}
\sum_{i \in \mathcal{N}_t} f_{\mathrm{S}}
\left(x^{(i)}_{t+1}\right) + \delta
\leq
\frac{1}{N_t}
\sum_{i \in \mathcal{N}_t} f
\left(x^{(i)}_{t+1}\right) + 2\,\delta.
\label{eqn-apply-it}
   \end{eqnarray}
Equation \eqref{eqn-apply-it} shows that we can relate the progress
of the server's global model $x_{t+1}$, in terms of the true loss $f$, with the local $f$-wise progress of clients' models
$x^{(i)}_{t+1}$. This is achieved in terms of \emph{honest clients} only:
 \eqref{eqn-apply-it} does not involve any attacked models
 $\tilde{x}^{(i)}_{t+1}$'s. Inequality \eqref{eqn-apply-it}
 can then be used, together with some advanced optimization analysis arguments, as a building block to derive convergence bounds under various settings, in terms of the clients' local update rules (e.g., SGD local steps), SGD noise assumptions, and losses $f$ and $f_{\mathrm{S}}$ function classes.

In what follows, we provide an experimental setup to validate our claim.

%% file: Experiments.tex
\section{Experiments}
\label{experiments}
\subsection{Experimental Setup}

We simulate a FL system comprising a central server and ten clients, focusing on image classification tasks using the CIFAR-10 \cite{krizhevsky2009learning}, FMNIST \cite{xiao2017fashion}, and MNIST \cite{yann2010mnist} datasets under IID data partitions. In the IID setting, the training dataset is randomly partitioned into ten equal subsets, each allocated to a distinct client. The server retains the test set associated with each dataset. 

To facilitate the proposed defense mechanism, the server's test set is split into two distinct subsets. The first, termed the \textit{evaluation} subset—equal in size to each client's local dataset—is used to assess the quality of the aggregated global model at the end of each communication round. The second subset is employed to compute the loss associated with each client's model updates, as detailed in Section~\ref{problem_setting}. In our experiments, the \textit{evaluation} subset is consistently used across all evaluated defense strategies to measure the prediction accuracy of the updated global model, defined as the ratio of correct predictions to the total number of predictions at the end of each communication round. For clarity and consistency throughout the experimental results, we refer to this evaluation metric as \textit{centralized accuracy}.

For the model architecture, we adopt two lightweight convolutional neural networks (CNNs) inspired by the respected implementations provided in the Flower federated learning framework repository. One network is designed specifically for the CIFAR-10 dataset \cite{krizhevsky2009learning}, while the other is used for both the MNIST \cite{yann2010mnist} and FMNIST \cite{xiao2017fashion} datasets. The machine learning task in the FL experiment is a multiclass classification problem involving 10 distinct classes across all evaluated datasets. Regarding local client-side training, we tune it by evaluating both conventional SGD and Adam optimizers using default parameter settings. Our experiments employ Adam, as it demonstrates marginally superior accuracy compared to SGD, confirming the findings of \cite{mills2021multi,reddi2020adaptive}, which emphasize Adam’s advantages in convergence speed and overall model performance relative to SGD.

Using the Flower framework, we conduct FL simulations spanning 50 communication rounds. Each simulation involves the evaluation of a specific defense mechanism under the influence of an adversarial attack. The defense strategy is applied from the beginning of the training process, while the adversarial attack is activated at the 15th round and persists throughout the remainder of the experiment. In all scenarios, 5 out of 10 participating clients are designated as malicious. The attack type remains consistent across all adversarial clients within each simulation and corresponds to those described in Section~\ref{Threat Models}. The examined defense mechanisms include the proposed approach alongside baseline techniques such as Mean, Trimmed Mean, Median, Krum, and Multi-Krum. An exception is the Bulyan defense, which requires \( c \geq 4m + 3 \), where \( c \) is the number of clients and \( m \) the number of Byzantine ones. Consequently, our configuration with high adversarial ratio (\( m = 5 \), \( c = 10 \)) violates this condition and is excluded from the experiments.

Most of the evaluated defense mechanisms require the initialization of their parameters. For the Trimmed Mean defense, the parameter $\beta$, which determines the fraction of extreme values (both highest and lowest) to be excluded as potential outliers, is set to the default value of $\beta=0.2$. In the cases of Krum and Multi-Krum, prior knowledge of the maximum number of adversarial clients $f$ is required; thus, $f$ is set to 5. Additionally, Multi-Krum requires specifying the number of client updates to retain for aggregation. This value is initialized to match the number of honest participants in the system.

The objective is to assess the resilience and responsiveness of each defense mechanism in maintaining stable and robust model training in the face of adversarial behavior aimed at degrading the global model’s accuracy. To this end, centralized accuracy is used as the key evaluation metric across all defense–attack pairings. Specifically, following the initiation of the adversarial attack at the 15th communication round in each simulation, the mean centralized accuracy of the updated global model is computed and examined, as detailed in the experimental results section. To support further exploration, the implementation used in our experiments is available in our public codebase\footnote{\url{https://github.com/Krith-man/WAFL25-RobustFL/}}.

\begin{table}[H]
\centering
\begin{tabular}{ll@{\hskip 10pt}c@{\hskip 10pt}c@{\hskip 10pt}c}
\toprule
\multirow{2}{*}{\textbf{Defense Scheme}} & \multirow{2}{*}{\hspace{7 pt}\textbf{Attack}} & \multicolumn{3}{c}{\textbf{Dataset}} \\
\cmidrule(lr){3-5}
& & \textbf{CIFAR-10} & \textbf{FMNIST} & \textbf{MNIST} \\
\midrule
\multirow{4}{*}{Loss-Based Clustering} 
  & No Attack      & $67.03 \pm 0.16$ & $88.9 \pm 0.48$ & $98.53 \pm 0.05$ \\
  & Label Flip     & $64.9 \pm 1.07$ & $\mathbf{88.67 \pm 0.25}$ & $\mathbf{98.41 \pm 0.08}$ \\
  & Sign Flip      & $\mathbf{65.48 \pm 0.28}$ & $\mathbf{88.9 \pm 0.16}$ & $\mathbf{98.41 \pm 0.06}$ \\
  & Gaussian Noise & $64.89 \pm 0.57$ & $\mathbf{88.86 \pm 0.27}$ & $\mathbf{98.34 \pm 0.04}$ \\
\midrule
\multirow{4}{*}{Mean} 
  & No Attack      & $67.31 \pm 0.42$ & $89.23 \pm 0.20$ & $98.43 \pm 0.04$ \\
  & Label Flip     & $31.78 \pm 1.34$ & $38.16 \pm 3.01$ & $36.71 \pm 0.43$ \\
  & Sign Flip      & $1.94 \pm 0.71$ & $10 \pm 0.00$ & $9.77 \pm 0.05$ \\
  & Gaussian Noise & $12.21 \pm 1.21$ & $32.76 \pm 5.97$ & $11.68 \pm 0.67$ \\
\midrule
\multirow{4}{*}{Trimmed Mean} 
  & No Attack      & $66.35 \pm 0.55$ & $89.23 \pm 0.1$ & $98.44 \pm 0.05$ \\
  & Label Flip     & $33.23 \pm 0.95$ & $37.16 \pm 4.15$ & $34.71 \pm 2.36$ \\
  & Sign Flip      & $1.68 \pm 0.15$ & $10 \pm 0.00$ & $10.1 \pm 0.31$ \\
  & Gaussian Noise & $13.46 \pm 1.22$ & $34.56 \pm 2.9$ & $12.42 \pm 1.07$ \\
\midrule
\multirow{4}{*}{Median} 
  & No Attack      & $66.37 \pm 0.64$ & $89.05 \pm 0.13$ & $98.44 \pm 0.04$ \\
  & Label Flip     & $33.22 \pm 0.24$ & $35.77 \pm 2.33$ & $41.25 \pm 1.03$ \\
  & Sign Flip      & $2.34 \pm 0.19$ & $10 \pm 0.00$ & $9.98 \pm 0.3$ \\
  & Gaussian Noise & $14.18 \pm 0.31$ & $34.51 \pm 0.64$ & $12.55 \pm 1.32$ \\
\midrule
\multirow{4}{*}{Krum} 
  & No Attack      & $60.31 \pm 0.23$ & $86.2 \pm 0.18$ & $97.57 \pm 0.06$ \\
  & Label Flip     & $60.18 \pm 0.71$ & $43.15 \pm 0.22$ & $48.69 \pm 0.07$ \\
  & Sign Flip      & $59.19 \pm 0.94$ & $86.29 \pm 0.43$ & $97.5 \pm 0.06$ \\
  & Gaussian Noise & $59 \pm 1.79$ & $86.44 \pm 0.13$ & $97.48 \pm 0.11$ \\
\midrule
\multirow{4}{*}{Multi Krum} 
  & No Attack      & $66.64 \pm 0.4$ & $89.14 \pm 0.14$ & $98.48 \pm 0.06$ \\
  & Label Flip     & $\mathbf{65.06 \pm 0.64}$ & $43.11 \pm 1.31$ & $49.82 \pm 1.4$ \\
  & Sign Flip      & $65.46 \pm 0.41$ & $88.75 \pm 0.06$ & $98.26 \pm 0.08$ \\
  & Gaussian Noise & $\mathbf{65.37 \pm 0.14}$ & $88.66 \pm 0.01$ & $98.31 \pm 0.17$ \\
\bottomrule
\end{tabular}

\caption{Centralized accuracy (\%) across defense schemes, attack types, and datasets in a 10-client FL setup with 50\% malicious participants. Each reported value represents the mean accuracy over 3 independent runs of the corresponding simulation. Bolded values indicate the highest accuracy for each dataset under a given attack scenario.}
\label{table:1}
\end{table}

\subsection{Experimental Results}
 
Table~\ref{table:1} reports the centralized accuracies obtained from all evaluated combinations of defense mechanisms and adversarial attack strategies. As demonstrated in the table, the proposed method consistently achieves superior performance compared to the \textit{Mean}, \textit{Trimmed Mean}, and \textit{Median} baselines across all datasets. This performance advantage stems from a key limitation of these traditional aggregation approaches: they either consider all client updates or apply simple statistical trimming, without the capacity to effectively discriminate between benign and adversarial contributions. Consequently, the global model parameters are adversely affected, resulting in suboptimal convergence and reduced learning efficacy.

\begin{figure}[]
    \centering
    \begin{subfigure}{\textwidth}
        \includegraphics[width=\linewidth]{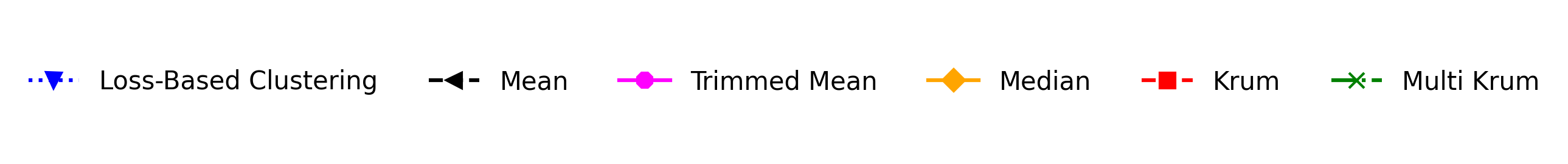}
    \end{subfigure}
    \begin{subfigure}{\textwidth}
        \includegraphics[width=0.9\linewidth]{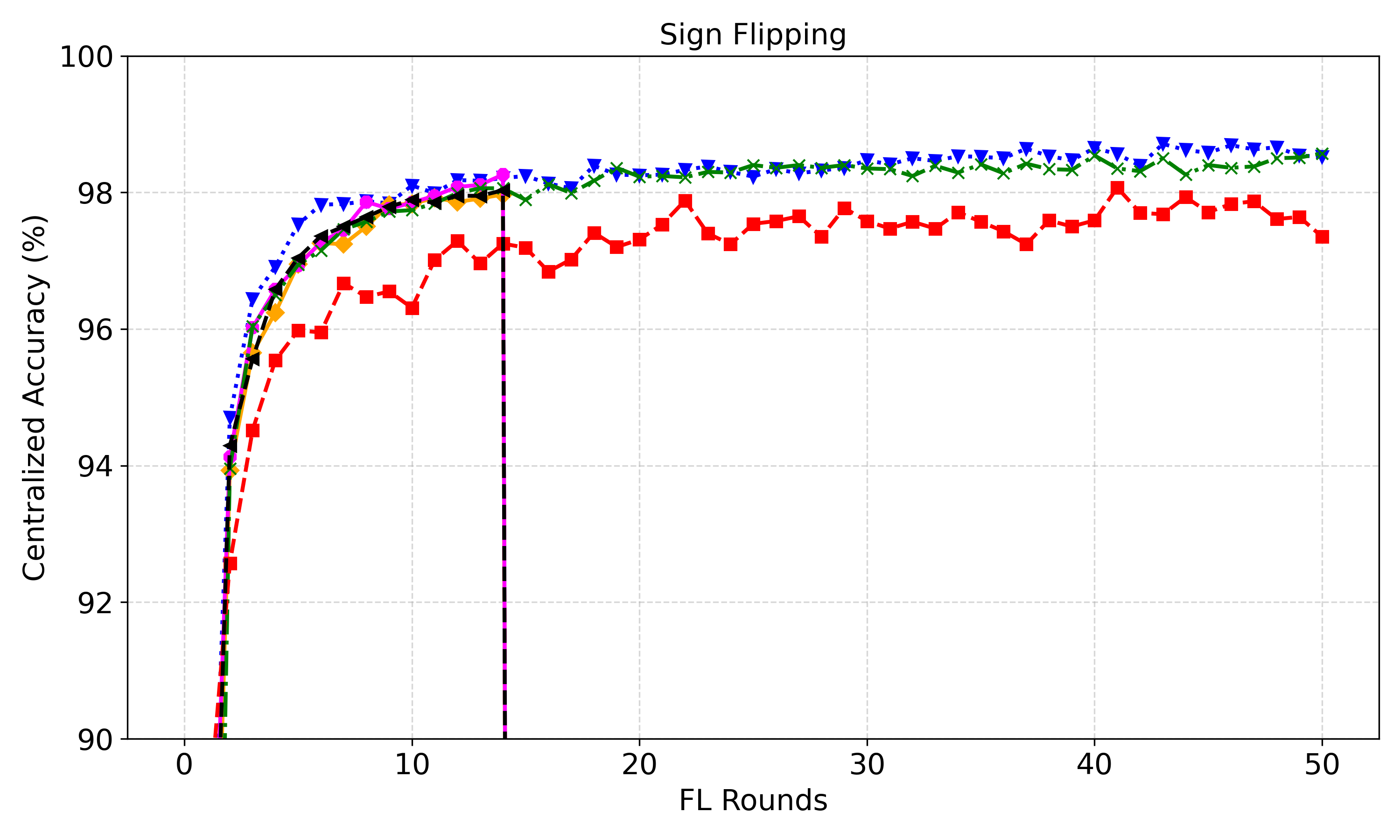}
    \end{subfigure}
    \begin{subfigure}{\textwidth}
        \includegraphics[width=0.9\linewidth]{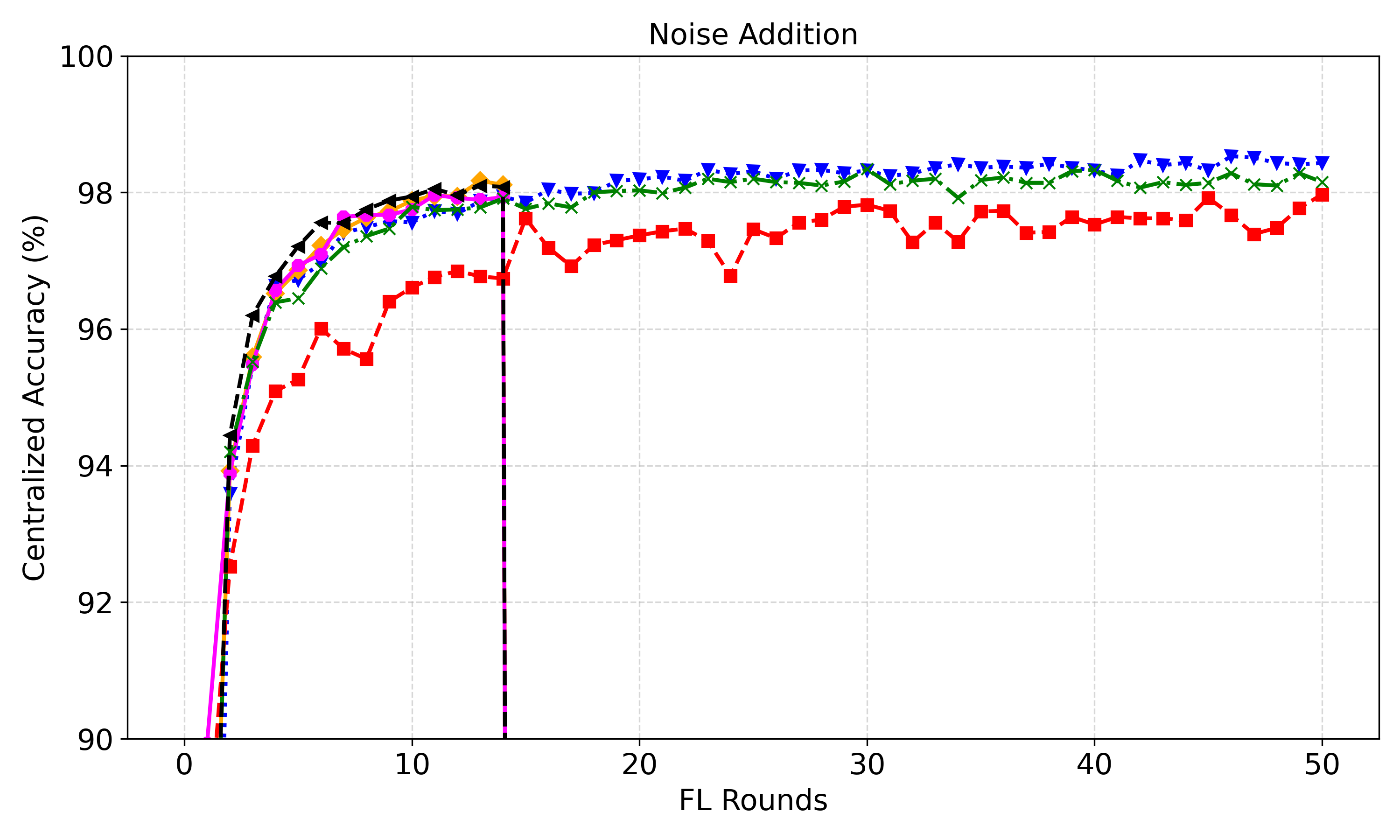}
    \end{subfigure}
    \begin{subfigure}{\textwidth}
         \includegraphics[width=0.9\linewidth]{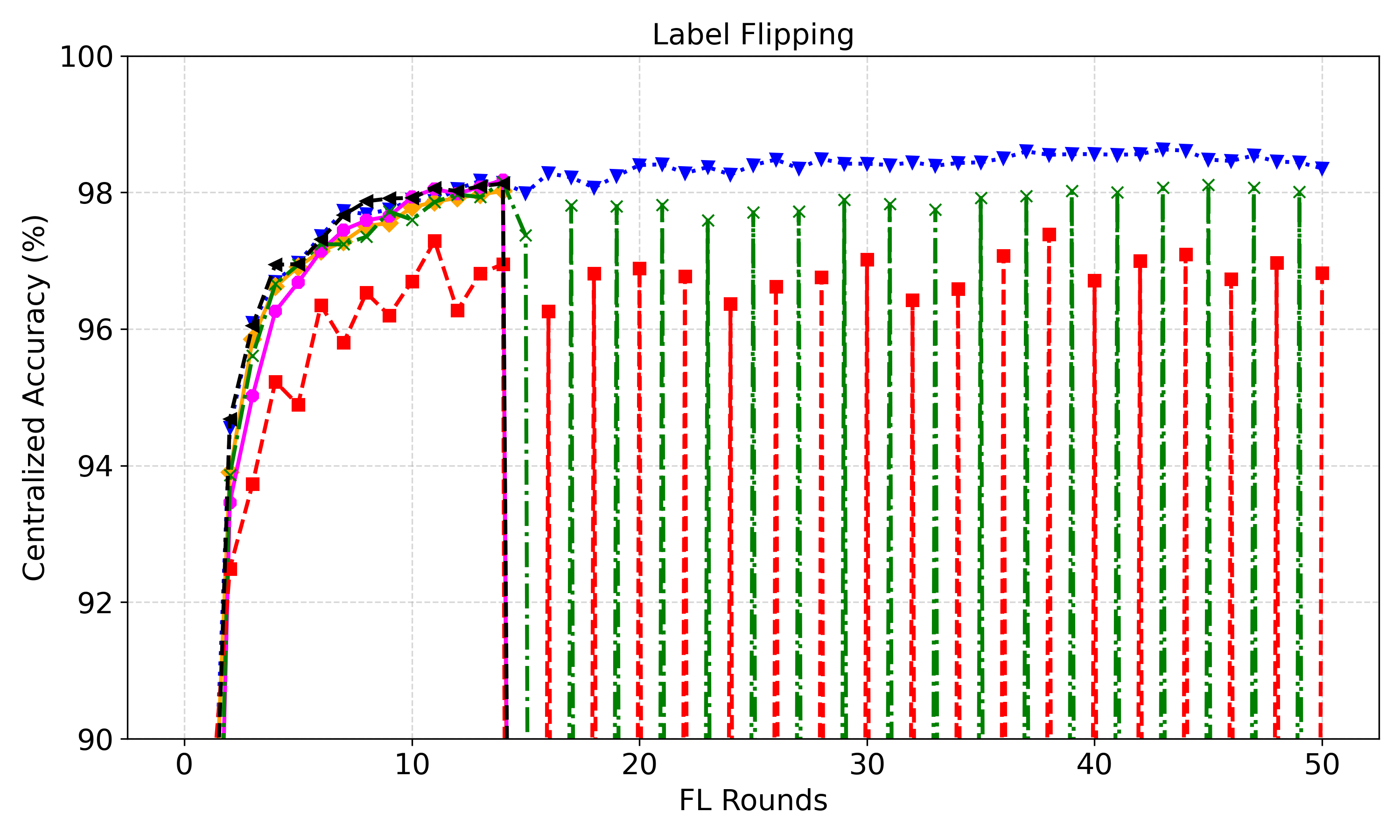}
    \end{subfigure}
    \caption{Centralized accuracy over 50 FL rounds on MNIST, comparing 6 defense methods against 3 distinct adversarial attacks in a 10-client FL setup with 50\% malicious participants.}
    \label{fig:1}
\end{figure}

This trend is further corroborated by the experimental results depicted in Figure~\ref{fig:1}, which illustrates the trajectory of centralized accuracy over 50 communication rounds for six defense methods under three different adversarial attack scenarios on the MNIST dataset. For clarity and brevity, we focus on illustrating the trajectory of centralized accuracy specifically for the MNIST case. The figure clearly shows that the baseline defenses suffer a sharp drop in accuracy following the onset of adversarial behavior, with no notable recovery observed throughout the training process. In contrast, the proposed defense demonstrates greater robustness and stability under adversarial conditions. 

Unlike \textit{Mean}, \textit{Trimmed Mean}, and \textit{Median} defense schemes—which aggregate model updates without discriminating between benign and malicious sources—the \textit{Krum} and \textit{Multi-Krum} algorithms selectively retain updates based on pairwise similarity, measured through Euclidean distance. These methods operate on the premise that updates from honest clients tend to be more similar to each other (i.e., exhibit smaller Euclidean distances) than those originating from adversarial clients. This property enables both \textit{Krum} and \textit{Multi-Krum} to effectively isolate and incorporate only trustworthy client updates in the aggregation process.

This assumption is empirically validated by the results, as the centralized accuracy achieved by \textit{Krum} remains close to that of the no-attack baseline across all datasets and attack types, indicating effective defense performance. However, our proposed method consistently surpasses \textit{Krum} in accuracy. This is attributed to a fundamental limitation in \textit{Krum}'s design: rather than aggregating updates from the selected subset of honest clients, it selects a single client update—specifically, the one with the minimal summed distance to others in the subset—as the new global model. Consequently, \textit{Krum} fails to leverage the full informative potential of honest client contributions. In contrast, our method incorporates aggregated information from multiple honest clients, resulting in improved model generalization and convergence. This advantage is clearly reflected in the performance gains shown in Table~\ref{table:1} and Figure~\ref{fig:1}.

To enhance the original \textit{Krum} defense mechanism, \textit{Multi-Krum} introduces an additional parameter that allows the user to specify the number of client updates - among those with the lowest \textit{Krum} scores—to be aggregated, rather than selecting only the single client update with the minimum score. This modification improves the centralized accuracy across all adversarial attack scenarios, yielding results comparable to those achieved by our proposed method.

However, both \textit{Krum} and \textit{Multi-Krum} exhibit two key limitations when compared to our approach. First, the computation of each client's \textit{Krum} score relies on identifying the $n - f - 2$ closest client updates, where $n$ denotes the total number of clients and $f$ represents the maximum number of assumed Byzantine participants. This means that both defense schemes require prior knowledge of the upper bound on malicious clients, whereas our method remains agnostic to this information. Second, the effectiveness of \textit{Krum}-based approaches depends heavily on the assumption that honest clients produce mutually similar model updates, forming a dense cluster in the parameter space. However, this assumption fails in attack scenarios where malicious updates are crafted to appear more similar to each other than the honest ones. A notable example is the Label-Flipping attack on the FMNIST and MNIST datasets, where the average centralized accuracy is reduced to nearly half of the No-Attack baseline. As shown in Figure~\ref{fig:1}, under the Label-Flipping attack of the MNIST dataset, both \textit{Krum} and \textit{Multi-Krum} exhibit unstable behavior, with centralized accuracy fluctuating sharply across rounds. This instability highlights their inability to consistently distinguish and select honest clients, particularly when adversarial updates are crafted to manipulate similarity metrics and confound the aggregation process.

%% file: Conclusion.tex
\section{Conclusions}
\label{conclusions}

In this work, we propose a client clustering approach based on model performance to distinguish between benign and potentially malicious updates, thereby enhancing the robustness of federated learning. The proposed approach dynamically filters out malicious contributions using client-side loss clustering. Our method effectively mitigates the impact of Byzantine attacks, including Gaussian noise addition, sign flipping, and data poisoning attacks, including label flipping. By comparing client-submitted model updates with their loss values on top of the server's trusted dataset, poisoned contributions can be effectively isolated, thereby promoting convergence. Experimental results on MNIST, FMNIST, and CIFAR-10 demonstrate that our approach outperforms standard and robust aggregation schemes—including Mean, Trimmed Mean, Median, Krum, and Multi-Krum—by achieving consistently higher centralized accuracy and stable convergence, even in scenarios with up to 50\% malicious clients. Importantly, the method does not require prior knowledge of the number of attackers and remains robust across diverse attack types, validating its practical effectiveness and generalization potential in decentralized FL settings. While our current experiments assume homogeneous data distributions, future work will explore the method’s effectiveness under heterogeneous data settings to broaden its applicability in real-world FL environments. Finally, we plan to investigate variants of the proposed method with client sampling, where only a subset of clients participate in each round.

%% file: Acknowledgements.tex
\subsubsection{\ackname} This paper has received funding from the European Union’s Horizon Europe research and innovation actions under grant agreement
No 101168560 (CoEvolution). Views and opinions expressed are however those of the author(s) only and do not necessarily reflect those of the European Union or the Commission. Neither the European Union nor the granting authority can be held responsible for them.
Also, it was supported by the European Union NextGenerationEU Recovery and Resilience Facility programme in the context of the RESEARCH-CREATE-INNOVATE 16971 framework (Project ATMO-MoRe, id: TAEDK-06199). The work of D. Jakovetic was supported by the Ministry of Science,
Technological Development and Innovation of the Republic of Serbia
(Grants No. 451-03-137/2025-03/200125 \& 451-03-136/2025-03/200125),
and by the Science Fund of the Republic of Serbia, Grant no. 7359,
project LASCADO.